\documentclass[letterpaper]{article} % DO NOT CHANGE THIS
\usepackage{aaai25}  % DO NOT CHANGE THIS
\usepackage{times}  % DO NOT CHANGE THIS
\usepackage{helvet}  % DO NOT CHANGE THIS
\usepackage{courier}  % DO NOT CHANGE THIS
\usepackage[hyphens]{url}  % DO NOT CHANGE THIS
\usepackage{graphicx} % DO NOT CHANGE THIS
\usepackage{natbib}  % DO NOT CHANGE THIS AND DO NOT ADD ANY OPTIONS TO IT
\usepackage{caption} % DO NOT CHANGE THIS AND DO NOT ADD ANY OPTIONS TO IT
\usepackage{algorithm}
\usepackage{algorithmic}

\usepackage[table,xcdraw]{xcolor}
\usepackage{tcolorbox}
\usepackage{times}
\usepackage{soul}
\usepackage{url}
\usepackage[utf8]{inputenc}
\usepackage{graphicx}
\usepackage{amsmath}
\usepackage{amsthm}
\usepackage{amsfonts}
\usepackage{booktabs}
\usepackage{algorithm}
\usepackage{algorithmic}
\usepackage[switch]{lineno}
\usepackage{enumitem}
\usepackage{multirow}
\usepackage{bm}
\usepackage{newfloat}
\usepackage{listings}
\DeclareCaptionStyle{ruled}{labelfont=normalfont,labelsep=colon,strut=off} % DO NOT CHANGE THIS
\floatstyle{ruled}
\newfloat{listing}{tb}{lst}{}
\floatname{listing}{Listing}
\title{DA-MoE: Addressing Depth-Sensitivity in Graph-Level Analysis \\ through Mixture of Experts}

\author {
         Zelin Yao\textsuperscript{\rm 1}\equalcontrib, Chuang Liu\textsuperscript{\rm 1}\equalcontrib, Xianke Meng\textsuperscript{\rm 1}\equalcontrib, Yibing Zhan\textsuperscript{\rm 2}, Jia Wu\textsuperscript{\rm 3}, \\ Shirui Pan\textsuperscript{\rm 4}, Wenbin Hu\textsuperscript{\rm 1}
}
\affiliations {
    \textsuperscript{\rm 1}School of Computer Science, Wuhan University
    \textsuperscript{\rm 2}JD Explore Academy 
    \textsuperscript{\rm 3}School of Computing, Macquarie University 
    \textsuperscript{\rm 4}School of Information and Communication Technology, Griffith University \\
    {zelinyao, chuangliu, xiankemeng, hwb}@whu.edu.cn, zybjy@mail.ustc.edu.cn, jia.wu@mq.edu.au, s.pan@griffith.edu.au
}

\usepackage{bibentry}
\begin{document}

\maketitle

\begin{abstract}
Graph neural networks (GNNs) are gaining popularity for processing graph-structured data. In real-world scenarios, graph data within the same dataset can vary significantly in scale. This variability leads to depth-sensitivity, where the optimal depth of GNN layers depends on the scale of the graph data. Empirically, fewer layers are sufficient for message passing in smaller graphs, while larger graphs typically require deeper networks to capture long-range dependencies and global features. However, existing methods generally use a fixed number of GNN layers to generate representations for all graphs, overlooking the depth-sensitivity issue in graph structure data. To address this challenge, we propose the depth adaptive mixture of expert (DA-MoE) method, which incorporates two main improvements to GNN backbone: \textbf{1)} DA-MoE employs different GNN layers, each considered an expert with its own parameters. Such a design allows the model to flexibly aggregate information at different scales, effectively addressing the depth-sensitivity issue in graph data. \textbf{2)} DA-MoE utilizes GNN to capture the structural information instead of the linear projections in the gating network. Thus, the gating network enables the model to capture complex patterns and dependencies within the data. By leveraging these improvements, each expert in DA-MoE specifically learns distinct graph patterns at different scales. Furthermore, comprehensive experiments on the TU dataset and open graph benchmark (OGB) have shown that DA-MoE consistently surpasses existing baselines on various tasks, including graph, node, and link-level analyses. The code are available at \url{https://github.com/Celin-Yao/DA-MoE}.
\end{abstract}

\section{Introduction}
\label{sec:introduction}

Graph neural networks (GNNs), renowned for their efficacy in capturing relationships by aggregating neighborhood information, have become increasingly popular in the graph-structured data field. 
In recent years, GNNs have been successfully applied in various areas, such as recommendation systems~\cite{fairsurvey,cola,collaborative}, social networks~\cite{gode,robust-gnn}, and molecular structures~\cite{pretrain-gnn,qh9}, achieving state-of-the-art performances in numerous tasks. 
Besides, GNNs can adjust their layers to aggregate information, making them adaptable to datasets with varying scales. However, even within a dataset, the scale of data can differ significantly. For instance, the number of nodes in the REDDIT-BINARY dataset can range from 6 to 3,782. Despite this variability, most existing approaches still use a fixed GNN layer configuration for all graphs, which may negatively impact the model’s performance by failing to consider the diverse scales of the graphs in the dataset.

\begin{figure}[!t] % !htb
\begin{center}
\includegraphics[width=0.9\linewidth]{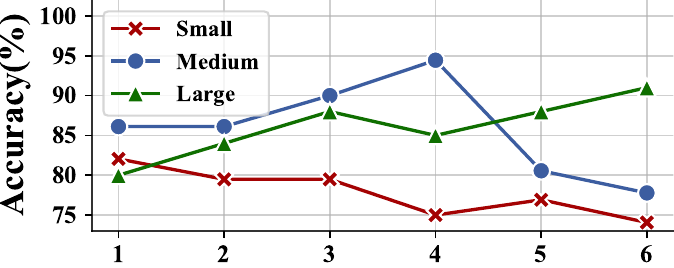}
\end{center}
\caption{The \textbf{depth-sensitivity} phenomenon in the IMDB-BINARY dataset, where graphs at different scales rely on specific GNN depths to capture information effectively.}
\label{fig:motiv}
\end{figure}

We conduct a series of experiments to further verify our observation. %what discussion,前面并没有一个明确的不同大小的graph需要不同深度的GNN，我建议可以用observation？或者哪些文章有提到这个。
Specifically, we categorized the graphs into three groups based on the number of nodes: graphs with fewer than 15 nodes are classified as small, those with 15 to 25 nodes as medium, and those with more than 25 nodes as large. Then, we split the dataset for the training and test sets, and report the performance on the test set. %实际上，给出1个曲线会更好，depth和size的大小分布。那么这个图可以1画出depth代表什么，scale代表什么，然后不同的scale对不同大小的性能。
Based on Figure~\ref{fig:motiv}, we identify a key insight: \textbf{Depth-Sensitivity:} We observe that shallow GNNs perform well on small-scale graphs, while deeper GNNs are suitable for large-scale graphs. We refer to the phenomenon where graphs of different scales require different GNN depths for effective aggregation as depth-sensitivity. In small-scale graphs, using deeper GNNs can lead to over-fitting and parameter redundancy. This increases the model's complexity and leads to inefficient and reduce performance. Conversely, large-scale graphs may struggle to learn global patterns or dependencies if the model suffers from under-reaching due to using a shallow set of GNN layers. Therefore, a fixed number of GNN layers may not be suitable for graphs of varying scales, limiting the model's generalization ability. Hence, an intuitive question raises: Is assigning an adaptive GNN layer model for different scale graphs within a dataset to capture their representations more effectively without any trial-and-error depth adjustments possible?
% \textbf{2) Time-consuming to determine GNN layers:} Typically, previous approaches manually select the optimal number of GNN layers by tedious trial-and-errors, seeking a balance point across different graphs to optimize overall performance. This approach is not only inefficient but also may not generalize well to new, unseen graphs, as the optimal number of layers could vary depending on the scale or specific characteristics of the graph.

% DA-MoE utilized the MoE module to automatically determine the optimal number of GNN layers for different scale graphs. 
To address this challenge, we propose the depth adaptive mixture of experts (DA-MoE) model. We introduce two main improvements into traditional GNN backbone. \textbf{1) MoE on GNN Layer:} DA-MoE addresses the depth-sensitivity issue by leveraging the MoE module to automatically determine the optimal number of GNN layers for different-scaled graphs. Specifically, this model utilizes GNNs of varying depths as experts, with each expert being an independent GNN model with its own parameters. For each graph, all experts generate their respective representations, and a gating network assigns scores to each expert. Moreover, graphs of different scales tend to select specific experts, enabling them to identify patterns and features unique to each scale. Subsequent visualization experiments of expert scores confirmed this hypothesis. \textbf{2) Structure-Based Gating Network:} Instead of utilizing linear projection to obtain the scores for each expert, our approach integrates structural information into the gating network. This modification adopts GNN as the gating network which aggregates information from neighbors, thereby improving the model's effectiveness and accuracy. By leveraging the structural information, our gating network can precisely allocate tasks to the most suitable experts, enhancing the model's overall performance. 

To validate DA-MoE's capability in addressing depth sensitivity, we conduct experiments on graph classification and regression tasks. Then, we extend DA-MoE to node and link-level tasks. These experiments included comparisons with the GNN backbone and other MoE methods across various datasets, including the TU dataset~\cite{tu-dataset} and the large-scale OGB dataset~\cite{ogb-dataset}.
The experimental results consistently indicate that DA-MoE surpasses state-of-the-art methods across the majority of datasets, with significant enhancements observed in the metrics. 
Our main contributions are summarized as follows:
\begin{enumerate}[leftmargin=12pt]
\item We introduce DA-MoE, a novel MoE framework that addresses the depth-sensitivity issue in graph-structured data. It adaptively assigns optimal GNN layers to graphs of different scales, enabling the model to flexibly aggregate information across various levels.
\item DA-MoE is a plug-and-play module that can be easily incorporated into existing GNN architectures. This design ensures that it can enhance the model's adaptability and efficiency without requiring significant modifications.
\item We conduct extensive experiments to compare DA-MoE with existing GNN and MoE models on 17 real-world graph datasets. The experimental results consistently validate the effectiveness of the proposed DA-MoE.
\end{enumerate}

\section{Related Work}
\label{sec:related work}

\paragraph{Graph Neural Networks.}
GNNs are designed to handle graph-structured data, achieving promising performance using a message passing mechanism to aggregate information from neighbors and update node representations. Traditional GNNs that achieve state-of-the-art performance in various tasks, including GatedGCN~\cite{gatedgcn}, GCN~\cite{gcn}, GAT~\cite{gat}, GIN~\cite{gin}, and GraphSAGE~\cite{graphsage}. Presently, researchers are focusing on improving GNN architecture to better aggregate graph information. For example, SAGIN~\cite{sagin} enhances GNNs by encoding subgraphs at different levels and inserting this information into nodes. Mao et al.\shortcite{disparity} proposed that the aggregation operation exhibits different effects on nodes with structural disparity. CF-GNN~\cite{cf-gnn} extends conformal prediction to traditional GNNs. Despite these advancements, existing GNN models often overlook the variations in data scale within a dataset, relying on a fixed layer configuration for all graphs. This approach can hinder their performance and reduce their adaptability when dealing with graphs of diverse scales and complexities.

\paragraph{Mixture of Expert Models.}
The MoE concept can be initially traced back to the work~\cite{moe,hi-moe}, which involved training a set of independent experts that specialized in learning distinct data features. Subsequently, a series of enhancement of MoE have been proposed, such as Aljundi et al.~\shortcite{expert-gate} introduced a model that determines the most pertinent expert network for a given task by employing an auto-encoder gate. Similarly, Noam Shazeer et al.~\shortcite{sparse-moe} presented a model that employs a sparse gating mechanism to activate only a subset of experts. Currently, the MoE module is widely applied in the computer vision~\cite{ramoe,moe-adapt} and natural language processing~\cite{switch,glam} domains. The integration of GNNs with MoE models is also advancing gradually. TopExpert~\cite{topexpert} utilizes a clustering-based gating module to categorize input molecules. GMoE~\cite{gmoe} incorporates multiple experts at each layer, enabling the model to learn different hop information. G-FAME~\cite{gfame} introduces a MoE module that leverages an ensemble of specialized neural networks to capture diverse facets of knowledge within the realm of fairness. Link-MoE~\cite{link-moe} utilize a range of existing linker predictors as experts, while GraphMETRO~\cite{graphmetro} introduces a novel mixture-of-aligned-experts architecture and training framework to address the distribution shift challenge.

\section{Preliminaries}
\label{sec:preliminary}

\paragraph{Notations.} A graph $\mathcal{G}(\boldsymbol{A}; \boldsymbol{X})$ consists an adjacency matrix $\boldsymbol{A} \in \{0, 1\}^{ n \times n}$ and a node feature matrix $\boldsymbol{X} \in \mathbb{R}^{ n \times d}$, where $n$ denotes the number of nodes, $d$ represents the node feature dimension, and $\boldsymbol{A}[i, j]=1$ indicates the presence of an edge between nodes $v_{i}$ and $v_{j}$, otherwise $\boldsymbol{A}[i, j]=0$.
\paragraph{Mixture of Experts.} The MoE~\cite{expert-gate,gmoe} comprises a collection of expert networks, denoted by $E = \{E_1, E_2, ..., E_n\}$, each with its own trainable parameters. Additionally, a gating network $Q$ is employed to score and select the experts based on their outputs. Given the initial feature $\boldsymbol{X}$, we denote $E(\boldsymbol{X}) = \{E(\boldsymbol{X})_i\}_{i=1}^s$ as experts' output and $Q(\boldsymbol{X}) = \{Q(\boldsymbol{X})_i\}_{i=1}^s$ as the gating network's, where $s$ is the number of experts. In addition, we normalize the scores to ensure stable training of the gating network and maintain expert diversity. Finally, the model output is derived by sequentially multiplying the outputs from the gating network with the representations generated by the expert models.

\begin{figure*}[!t] % !htb
\begin{center}
\includegraphics[width=0.95\linewidth]{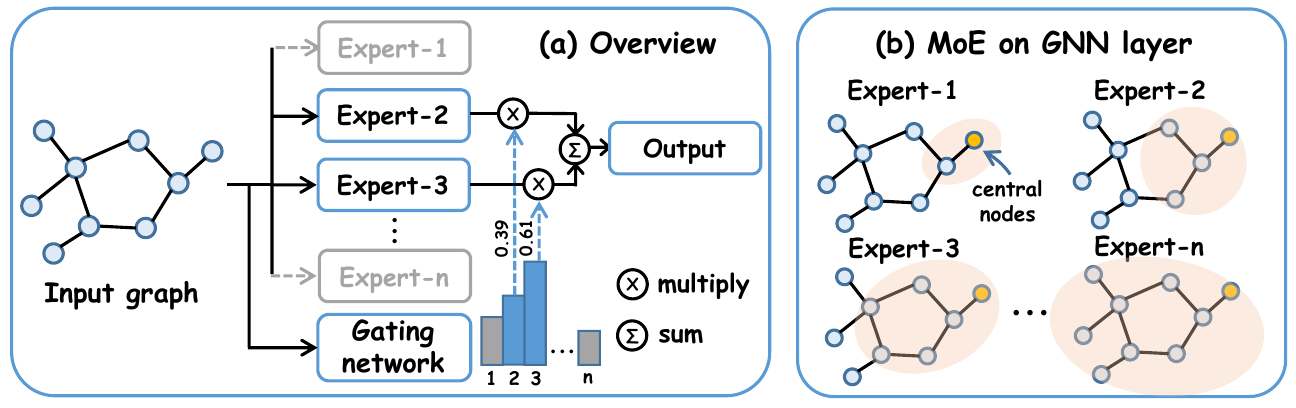}
\end{center}
\caption{Overview architecture of DA-MoE. (a) We utilize multiple experts instead of GNN backbone to learn specialized patterns from various aggregation scales. The gating network activates only a subset of these experts, with the grey arrows denoting the experts that were not activated. (b) Each expert is a GNN network with a distinct layer, where ``Expert-1" refers to a 1-layer GNN. The shaded area in the figure represents the information aggregation range of the central node.
}
\label{fig:model}
\end{figure*}

\section{DA-MoE: Proposed Method}
\label{sec:method}
% In this section, we introduce the framework of DA-MoE, as illustrated in Figure~\ref{fig:model}. First, we introduce the DA-MoE layer which integrate MoE with GNN layer to solve depth-sensitivity phenomenon in graph data. Then, we propose structure based gating network. In addition, we adopt balance loss function to prevent the gating network always select the single group of experts. Next, we introduce the general framework of DA-MoE. Finally, we analyzed the efficiency of DA-MoE.

\subsection{MoE on GNN layer}
\label{sec:moe-gnn}
In graph-structured data, the GNN layer depth determines the scale of the neighborhood information that can be captured. Existing methods employ a fixed layer configuration to graphs of varying scales within a dataset. However, using a fixed number of GNN layers results in under-reaching for large graphs, while causing over-fitting for small ones, neglecting depth-sensitivity. %建议在哪里要把这个定义清楚
Thus, we integrate the MoE module with the GNN layer, treating GNNs at different layers as experts. This approach enables adaptive learning and efficiently captures the varying scale of neighborhood information across different graphs. Given the $i$-th expert $E(\boldsymbol{X})_i$ is a $L$-th layer GNN, we can formulate the following:
\begin{equation}
\begin{aligned}
    \boldsymbol{H}_v^{(l)} &= \mathrm{GNN}^{(l)}(\boldsymbol{H}_v^{(l-1)}, \boldsymbol{A}), \\
    E(\boldsymbol{X})_i &=  \mathrm{READOUT}(\{\boldsymbol{H}_v^{(L)}|v \in \mathcal{V}\}),
\end{aligned}
\end{equation}
where $\mathrm{GNN}^{(l)}(\cdot)$ is the $l$-th layer of the GNN backbone, with $\boldsymbol{H}_v^{(0)} = \boldsymbol{X}_v \in \mathbb{R}^d$, $\mathrm{READOUT}(\cdot)$ function aggregates node features from the final iteration to obtain the entire graph’s representation, and $E(\cdot)_i$ is the $i$-th expert output. By adopting distinct aggregation scales, each expert is capable of capturing distinct aspects of knowledge and patterns, thereby enhancing the overall performance of the model.

In graph-level tasks, the depth-sensitivity phenomenon refers to graph representations being easily influenced by the graph's scale. Similar to depth-sensitivity in graph-level tasks, node representations are also susceptible to the influence of node degrees during aggregation. High-degree nodes may receive excessive information, which introduces noise, while low-degree nodes might not gather enough information at the same scale. Due to the varying degrees within a graph, fixing the aggregation scale for a node can lead to suboptimal performance. To address this issue, DA-MoE allows each node to select a specific expert, enabling the model to dynamically adjust the aggregation scales. Unlike Eq. (1), the experts' outputs are directly obtained from the GNN layer without applying the $\mathrm{READOUT}$ function. Thus, these modifications enables DA-MoE to be easily applied to node and link-level tasks, which require node representation for classification or prediction.

\subsection{Structure-Based Gating Network}
\label{sec:gate}

The gating network is a central component of the MoE module. By dynamically selecting a subset of experts, it enables the model to focus on the most relevant experts for a given task, thereby enhancing overall efficiency and performance. In the study~\cite{expert-gate,gmoe} linear projection was employed to obtain decision scores. Specifically, they utilized the noise top-k gating strategy as the score function. This strategy introduces two key characteristics: sparsity and noise. Sparsity enables the activation of only a subset of experts, which conserves computational resources. Similarly, noise helps stabilize the model and prevents it from consistently selecting the same experts. However, linear projection based gating networks ignore the graph's structural information, limiting their ability to capture relationships and patterns. Thus, we propose the structure-based gating network, which incorporates topological relationships and neighborhood information. This can be described as:
\begin{equation}
\begin{aligned}
    T(\boldsymbol{X}) &= \sigma \left((1+\alpha) \cdot \boldsymbol{X}_v + \sum\nolimits_{u \in \mathcal{N}(v)} \boldsymbol{X}_u \right), \\
    U(\boldsymbol{X}) &= T(\boldsymbol{X}) + z \cdot \mathrm{log}\left(1+\mathrm{exp}\left(T\left(\boldsymbol{X}\right)\right)\right), \\
    Q(\boldsymbol{X}) &= \mathrm{Softmax}\left(\mathrm{TopK}\left(U\left(\boldsymbol{X}\right),k\right)\right),
\end{aligned}
\end{equation}
where $\boldsymbol{X}_v$ denotes the representation of nodes $v$, $\alpha$ refers to a learnable parameter adjusting the contribution of the node's own feature, $\sigma(\cdot)$ denotes a two-layer fully connected neural network with a nonlinear activation function applied between the layers, and $\mathcal{N}(v)$ is a set of 1-hop neighborhood for node $v$.  $U(\cdot)$ includes the sparsity and noise characteristics and utilizes GNN to capture the graph's structural information while scoring. %这个地方的GNN有讲究么》
In addition, $z \in \mathcal{N}(0,1)$ indicates the standard Gaussian distribution, which controls the level of noise introduced into the model. Finally, $\mathrm{TopK}(U(\boldsymbol{X}), k)$ returns the index of the largest $k$ values of $U(\boldsymbol{X})$, introducing expert sparsity. 

\subsection{Balanced Loss Function}
\label{sec:loss}
In the MoE module, we discovered a common issue known as mode collapse. This phenomenon is characterized by the gating network consistently selecting a few experts, leading to an imbalance during the training process. Consequently, the frequently chosen experts have high scores, leading to further preferential selection. To address these issues, we introduced two additional balanced loss functions following these works~\cite{gmoe,expert-gate}. The first loss function is designed to resolve experts' imbalanced scores, and the second the unequal probabilities problem that occurs during expert selection. %这个是独特提出来的么？还是之前有的
The first loss function can be expressed as:
\begin{equation}
    CV(\boldsymbol{S})^2 = \frac{\frac{1}{N_d} \sum_{i=1}^{N_d} (\boldsymbol{S}_i - \mu)^2}{\mu^2+\epsilon},
\end{equation}
\begin{equation}
     \mathcal{L}_1(\boldsymbol{X}) = CV\left(\sum_{\boldsymbol{x} \in \boldsymbol{X}} Q(\boldsymbol{x})\right)^2, 
\end{equation}
where $CV(\cdot)$ denotes the coefficient of variation, which encourages a more uniform positive distribution, $\mu$ denotes the mean tensor of $\boldsymbol{S}$, and $\epsilon$ represents a small positive constant added for numerical stability. Though the $\mathcal{L}_1$ loss function can ensure that the scores given to each expert are equal, situations where a single expert is assigned scores to various graphs can still occur. Besides, the scores of experts are discrete, making it challenging to achieve a balanced scores among the experts. Thus, we implement a load balancing loss to ensure that the probability of each expert being selected is equal. We define $P(\boldsymbol{X}, i)$ as the probability of $Q(\boldsymbol{X})\neq 0$, which can be written as:
\begin{equation}
    P(\boldsymbol{X}, i) = Pr(U(\boldsymbol{X})_i > \mathrm{TopK\_ex}(U(\boldsymbol{X}),k,i)),
\end{equation}
where $\mathrm{TopK\_ex}(\cdot,k,i)$ denotes the $k$ largest elements while excluding $i$-th element. Thus, the loss function is defined as:
\begin{equation}
    \mathcal{L}_2(\boldsymbol{X}) = CV\left(\sum_{\boldsymbol{x} \in \boldsymbol{X}} P(\boldsymbol{x},i)\right)^2,
\end{equation}
Finally, the model's final loss is computed by adding two additional balanced loss functions:
\begin{equation}
    \mathcal{L}_{final} = \mathcal{L} + \lambda_1 \cdot \mathcal{L}_1 + \lambda_2 \cdot \mathcal{L}_2,
\end{equation}
where $\mathcal{L}$ is cross entropy loss for the classification task and mean square error loss for the regression task, and $\lambda_1$ and $\lambda_2$ are the factors for the two balanced losses, respectively.
\subsection{General Framework of DA-MoE}
\label{sec:moe-model}
The general DA-MoE framework is illustrated in Figure~\ref{fig:model}. DA-MoE can be applied to any GNNs by substituting each layer with a plug-and-play DA-MoE layer. In the DA-MoE layer, the graph's features are sent to the gating network to produce normalized scores $Q(\boldsymbol{X})_i$ for each expert. These scores indicate each expert's contribution to the final embeddings. Meanwhile, each expert processes the graph information and generates an embedding $E(\boldsymbol{X})_i$. The final embedding is calculated by summing the weighted scores, with top-k experts being activated. Specifically, the DA-MoE framework's output embedding can be expressed as:
\begin{equation}
    \boldsymbol{h}_o = \sum_{i \in \mathcal{A}} Q(\boldsymbol{X})_iE(\boldsymbol{X})_i,
\end{equation}
where $\mathcal{A}$ indicates the subset of experts selected from the top-k function as defined in Eq. (2). In summary, the DA-MoE model's main advantage is dynamically aggregating and propagating information from neighbors at varying distances, effectively addressing the depth-sensitivity issue. 
\subsection{Complexity Analysis}
\label{sec:time_analysis}
In this section, we analyze the DA-MoE model's computational complexity. As previously discussed, we replaced the traditional GNN layer with the DA-MoE one. The complexity of a $l$-th GNN layer can be represented as $\mathcal{O}(l \cdot (|\mathcal{E}|d + nd^2))$, with $|\mathcal{E}|$, $n$, and $d$ referring to number of edges, number of nodes, and hidden dimension respectively. Since our DA-MoE model incorporates MoE modules into the GNN layers, let $K$ represent the number of GNN layers of the selected experts. During the inference process, the complexity is $\mathcal{O}(\sum_{k \in K} k \cdot (|\mathcal{E}|d + nd^2)))$, which is related to the number of selected experts. Notably, when $k=1$, the DA-MoE model's complexity is approximate to a typical fixed-layer GNN. The detailed experimental results are presented in the experiment section.
\begin{table*}[!t]
\centering
\renewcommand\arraystretch{1.25} % 行间距
\setlength\tabcolsep{15pt} % 列间距
\resizebox{\textwidth}{!}{%
\begin{tabular}{@{}lccccccc@{}}
\toprule
\multirow{2}{*}{} &
  \multicolumn{3}{c}{\textbf{Biochemical Domain}} &
  \multicolumn{4}{c}{\textbf{Social Domain}} \\ \cmidrule(l){2-4} \cmidrule(l){5-8}
 &
  PROTEINS &
  NCI1 &
  MUTAG &
  IMDB-B &
  IMDB-M &
  COLLAB &
  RDT-B \\ \midrule
\multicolumn{1}{l}{GIN} &
  $79.91_{\pm 0.62}$ &
  $83.28_{\pm 0.33}$ &
  $95.39_{\pm 0.75}$ &
  $77.77_{\pm 0.59}$ &
  $53.63_{\pm 0.36}$ &
  $82.59_{\pm 0.21}$ &
  $84.24_{\pm 1.75}$ \\
\multicolumn{1}{l}{\quad + DA-MoE} &
  $\textbf{80.27}_{\pm \textbf{0.39}}$ &
  $\textbf{83.57}_{\pm \textbf{0.27}}$ &
  $\textbf{95.93}_{\pm \textbf{0.57}}$ &
  $\textbf{78.78}_{\pm \textbf{0.50}}$ &
  $\textbf{55.09}_{\pm \textbf{0.20}}$ &
  $\textbf{83.89}_{\pm \textbf{0.19}}$ &
  $\textbf{84.72}_{\pm \textbf{1.79}}$ \\
\multicolumn{1}{l}{\quad \textbf{Improvement}} &
  $0.45 \%$ &
  $0.35 \%$ &
  $0.57 \%$ &
  $1.30 \%$ &
  $2.72 \%$ &
  $1.57 \%$ &
  $0.57 \%$ \\ \midrule
\multicolumn{1}{l}{GCN} &
  $78.68_{\pm 0.37}$ &
  $75.94_{\pm 0.25}$ &
  $87.50_{\pm 0.57}$ &
  $\textbf{80.13}_{\pm \textbf{0.31}}$ &
  $55.25_{\pm 0.32}$ &
  $84.18_{\pm 0.19}$ &
  $86.40_{\pm 1.45}$ \\
\multicolumn{1}{l}{\quad + DA-MoE} &
  $\textbf{78.99}_{\pm \textbf{0.39}}$ &
  $\textbf{80.63}_{\pm \textbf{0.44}}$ &
  $\textbf{93.00}_{\pm \textbf{1.03}}$ &
  $79.92_{\pm 0.39}$ &
  $\textbf{55.95}_{\pm \textbf{0.18}}$ &
  $\textbf{84.62}_{\pm \textbf{0.38}}$ &
  $\textbf{87.07}_{\pm \textbf{0.86}}$ \\
\multicolumn{1}{l}{\quad \textbf{Improvement}} &
  $0.39 \%$ &
  $6.18 \%$ &
  $6.29 \%$ &
  -- &
  $1.27 \%$ &
  $0.52 \%$ &
  $0.78 \%$ \\ \midrule
\multicolumn{1}{l}{GatedGCN} &
  $79.22_{\pm 0.25}$ &
  $\textbf{80.24}_{\pm \textbf{0.39}}$ &
  $92.11_{\pm 1.05}$ &
  $78.02_{\pm 0.61}$ &
  $53.33_{\pm 0.40}$ &
  $76.23_{\pm 0.96}$ &
  $81.69_{\pm 0.99}$ \\
\multicolumn{1}{l}{\quad + DA-MoE} &
  $\textbf{79.77}_{\pm \textbf{0.43}}$ &
  $78.40_{\pm 0.48}$ &
  $\textbf{93.56}_{\pm \textbf{0.90}}$ &
  $\textbf{78.13}_{\pm \textbf{0.46}}$ &
  $\textbf{54.78}_{\pm \textbf{0.37}}$ &
  $\textbf{83.25}_{\pm \textbf{0.10}}$ &
  $\textbf{82.23}_{\pm \textbf{1.27}}$ \\
\multicolumn{1}{l}{\quad \textbf{Improvement}} &
  $0.69 \%$ &
  -- &
  $1.57 \%$ &
  $0.14 \%$ &
  $2.72 \%$ &
  $9.21 \%$ &
  $0.66 \%$ \\ \bottomrule
\end{tabular}%
}
\caption{\textbf{Graph classification} experiments results on seven datasets from TUDataset with biochemical and social domain. The reported metrics are mean accuracy and standard deviation, the higher the better. The most outstanding results are highlighted in \textbf{bold}, and the improvements achieved by DA-MoE are listed below each dataset.}
% Symbol `+' signifies the incorporation of the module into the backbone GNN.
\label{tab:result-tu}
\end{table*}

\definecolor{lightpink}{HTML}{FAEAE1}
\definecolor{iceblue}{HTML}{E0F5FF}
\tcbset{
  pinkbox/.style={
    colback=lightpink,
    colframe=lightpink,
    width = 1.3cm,
    height = 0.35cm,
    halign=center, % 居中对齐
    valign=center, % 竖直居中
  }
}
\tcbset{
  bluebox/.style={
    colback=iceblue,
    colframe=iceblue,
    width = 1.3cm,
    height = 0.35cm,
    halign=center, % 居中对齐
    valign=center, % 竖直居中
  }
}

\begin{table*}[!t]
\centering
\renewcommand\arraystretch{1.50} % 行间距
\setlength\tabcolsep{4pt} % 列间距
\resizebox{\textwidth}{!}{%
\begin{tabular}{@{}lcccccc@{}}
\toprule
\multirow{2}{*}{} & \multicolumn{4}{c}{\textbf{OGB Graph Classification} (ROC-AUC)}                    & \multicolumn{2}{c}{\textbf{OGB Graph Regression} (RMSE)} \\ \cmidrule(l){2-5} \cmidrule(l){6-7}
                  & \textbf{ogbg-molhiv}              & \textbf{ogbg-moltox21}            & \textbf{ogbg-moltoxcast}          & \textbf{ogbg-molbbbp}             & \textbf{ogbg-molesol}                    & \textbf{ogbg-molfreesolv}              \\ \midrule
GIN               & $75.40_{\pm 1.50} $   & $74.30_{\pm 0.50}$   & $63.30_{\pm 1.50}$   & $65.50_{\pm 1.80}$   & $1.17_{\pm 0.06}$      & $2.76_{\pm 0.35}$     \\ \cmidrule(lr){2-2} \cmidrule(lr){3-3} \cmidrule(lr){4-4}  \cmidrule(lr){5-5} \cmidrule(lr){6-6}  \cmidrule(lr){7-7} 
\quad + GMoE             
& $76.14_{\pm 1.03} \ \footnotesize \begin{tcolorbox}[bluebox, left=-1.5pt, baseline=0.1cm]{\text{$\uparrow 0.98\%$}} \end{tcolorbox}$ 
& $74.76_{\pm 0.66} \ \footnotesize \begin{tcolorbox}[bluebox, left=-1.5pt, baseline=0.1cm]{\text{$\uparrow 0.62\%$}} \end{tcolorbox}$ 
& $62.86_{\pm 0.37} \ \footnotesize \begin{tcolorbox}[bluebox, left=-1.5pt, baseline=0.1cm]{\text{$\downarrow 0.70\%$}} \end{tcolorbox}$ 
& $66.93_{\pm 1.72} \ \footnotesize \begin{tcolorbox}[bluebox, left=-1.5pt, baseline=0.1cm]{\text{$\uparrow 2.18\%$}} \end{tcolorbox}$ 
& $1.20_{\pm 0.03} \ \footnotesize \begin{tcolorbox}[bluebox, left=-1.5pt, baseline=0.1cm]{\text{$\downarrow 2.50\%$}} \end{tcolorbox}$      
& $2.56_{\pm 0.24} \ \footnotesize \begin{tcolorbox}[bluebox, left=-1.5pt, baseline=0.1cm]{\text{$\uparrow 7.25\%$}} \end{tcolorbox}$     \\
\quad + DA-MoE         
& $\textbf{77.47}_{\pm \textbf{1.77}} \ \footnotesize \begin{tcolorbox}[pinkbox, left=-1.5pt, baseline=0.1cm]{\text{$\uparrow \textbf{2.75\%}$}} \end{tcolorbox}$ 
& $\textbf{75.33}_{\pm \textbf{0.73}} \ \footnotesize \begin{tcolorbox}[pinkbox, left=-1.5pt, baseline=0.1cm]{\text{$\uparrow \textbf{1.39\%}$}} \end{tcolorbox}$ 
& $\textbf{64.55}_{\pm \textbf{0.46}} \ \footnotesize \begin{tcolorbox}[pinkbox, left=-1.5pt, baseline=0.1cm]{\text{$\uparrow \textbf{1.97\%}$}} \end{tcolorbox}$ 
& $\textbf{67.88}_{\pm \textbf{1.58}} \ \footnotesize \begin{tcolorbox}[pinkbox, left=-1.5pt, baseline=0.1cm]{\text{$\uparrow \textbf{3.63\%}$}} \end{tcolorbox}$ 
& $\textbf{1.14}_{\pm \textbf{0.40}} \ \footnotesize \begin{tcolorbox}[pinkbox, left=-1.5pt, baseline=0.1cm]{\text{$\uparrow \textbf{2.56\%}$}} \end{tcolorbox}$      
& $\textbf{2.51}_{\pm \textbf{0.19}} \ \footnotesize \begin{tcolorbox}[pinkbox, left=-1.5pt, baseline=0.1cm]{\text{$\uparrow \textbf{9.06\%}$}} \end{tcolorbox}$     \\ \midrule
GCN              
& $76.06_{\pm 0.97}$ & $75.29_{\pm 0.69}$ & $63.54_{\pm 0.42}$ & $68.87_{\pm 1.51}$ & $1.11_{\pm 0.04}$       & $2.64_{\pm 0.24}$      \\ \cmidrule(lr){2-2} \cmidrule(lr){3-3} \cmidrule(lr){4-4}  \cmidrule(lr){5-5} \cmidrule(lr){6-6}  \cmidrule(lr){7-7} 
\quad + GMoE            
& $77.35_{\pm 0.63} \ \footnotesize \begin{tcolorbox}[bluebox, left=-1.5pt, baseline=0.1cm]{\text{$\uparrow 1.70\%$}} \end{tcolorbox}$ 
& $75.45_{\pm 0.58} \ \footnotesize \begin{tcolorbox}[bluebox, left=-1.5pt, baseline=0.1cm]{\text{$\uparrow 0.21\%$}} \end{tcolorbox}$ 
& $64.12_{\pm 0.61} \ \footnotesize \begin{tcolorbox}[bluebox, left=-1.5pt, baseline=0.1cm]{\text{$\uparrow 0.91\%$}} \end{tcolorbox}$ 
& $\textbf{70.04}_{\pm \textbf{1.12}} \ \footnotesize \begin{tcolorbox}[bluebox, left=-1.5pt, baseline=0.1cm]{\text{$\uparrow \textbf{1.70\%}$}} \end{tcolorbox}$ 
& $\textbf{1.09}_{\pm \textbf{0.04}} \ \footnotesize \begin{tcolorbox}[bluebox, left=-1.5pt, baseline=0.1cm]{\text{$\uparrow \textbf{1.80\%}$}} \end{tcolorbox}$       
& $2.50_{\pm 0.19}$  \ \footnotesize \begin{tcolorbox}[bluebox, left=-1.5pt, baseline=0.1cm]{\text{$\uparrow 5.30\%$}} \end{tcolorbox}    \\
\quad + DA-MoE 
& $\textbf{77.62}_{\pm \textbf{1.58}} \ \footnotesize \begin{tcolorbox}[pinkbox, left=-1.5pt, baseline=0.1cm]{\text{$\uparrow \textbf{2.05\%}$}} \end{tcolorbox}$ 
& $\textbf{75.59}_{\pm \textbf{0.69}} \ \footnotesize \begin{tcolorbox}[pinkbox, left=-1.5pt, baseline=0.1cm]{\text{$\uparrow \textbf{0.40\%}$}} \end{tcolorbox}$ 
& $\textbf{65.18}_{\pm \textbf{0.48}} \ \footnotesize \begin{tcolorbox}[pinkbox, left=-1.5pt, baseline=0.1cm]{\text{$\uparrow \textbf{2.58\%}$}} \end{tcolorbox}$ 
& $69.62_{\pm 0.98} \ \footnotesize \begin{tcolorbox}[pinkbox, left=-1.5pt, baseline=0.1cm]{\text{$\uparrow 1.52\%$}} \end{tcolorbox}$ 
& $1.13_{\pm 0.04} \ \footnotesize \begin{tcolorbox}[pinkbox, left=-1.5pt, baseline=0.1cm]{\text{$\downarrow 1.80\%$}} \end{tcolorbox}$ 
& $\textbf{2.19}_{\pm \textbf{0.07}} \ \footnotesize \begin{tcolorbox}[pinkbox, left=-2.5pt, baseline=0.1cm]{\text{$\uparrow \textbf{17.05\%}$}} \end{tcolorbox}$ \\ \bottomrule
\end{tabular}
}
\caption{\textbf{OGB graph classification and regression} experiments on six datasets. The metrics are ROC-AUC and RMSE scores for classification and regression tasks respectively. For ROC-AUC, higher scores indicate better performance, while for RMSE, lower scores are preferred. The upward arrow ($\uparrow$) indicates an improvement in model performance compared to the GNN backbone, while the downward arrow ($\downarrow$) signifies a decrease in performance for each respective backbone.}
\label{tab:result-ogb}
\end{table*}
\section{Experiment}
\label{sec:experiment}
In this section, we conduct the experiment on graph-level task to demonstrate that DA-MoE can effectively solve the depth-sensitivity issue. In the process of propagating node information, there is a similar issue that the degrees of the nodes can influence the nodes representation. Then, we extend this framework to node and link-level tasks. In addition, we conduct a series of experiments to validate the effectiveness of DA-MoE from various aspects. 
\subsection{Experimental Settings}
\label{sec:settings}
\paragraph{Baselines.} To demonstrate the effectiveness of our proposed method, we compared DA-MoE with three GNN backbone models, including GCN~\cite{gcn}, GIN~\cite{gin}, and GatedGCN~\cite{gatedgcn}. We also used popular MoE model like GMoE~\cite{gmoe}. For all the datasets and tasks, we leveraged results from previous studies wherever possible. In addition, for datasets where baseline results are not provided, we conducted the experiments using the same parameters as our model. The detailed descriptions of all baseline models and configurations can be found in the Appendix.
% \paragraph{Dataset Split.} For TUDataset~\cite{tu-dataset}, we adopted a 10-fold-cross-validation approach, where the dataset is divided into training and test sets in an 9:1 ratio. Then, we repeated the experiment with 10 different random seeds to obtain the mean performance. For Open Graph Benchmark (OGB) dataset~\cite{ogb-dataset}, which has already defined the scaffold split with the split ratio of 80/10/10, and we conduct 10 trials with random seeds.

% we use Adam~\cite{adam} optimizer with $\beta_1 = 0.9$, $\beta_2 = 0.999$, $\epsilon = 1e-8$ and weight decay is set to $5 \times 10^{-4}$. Then,

% For other hyper-parameters, we grid searched the scaling factor $\lambda_1=\lambda_2$ in the set $\{0, 0.0001, 0.001, 0.01, 0.1, 1\}$ and the number of experts in the set $\{1, 2, 3, 4, 5\}$, $k$ in the set $\{1, 2, 3, 4, 5\}$.
\paragraph{Implementation Details.}
We assessed DA-MoE's effectiveness by measuring its performance in all tasks using supervised signals. For all the experiments, we follow the official implement in previous works~\cite{gmoe} and maintain the same model configurations, facilitating a direct evaluation of DA-MoE's performance relative to established benchmarks. To ensure reliability, we conduct 10 trials for each model using different random seeds and report the mean and deviation of the results. For the graph-level task, each graph can select embeddings from different aggregation scales as its output. For the node and link-level task, each node is allowed to adaptively select the experts. All the experiments were conducted on a Linux server with two NVIDIA A100s with 40GB memory. Detailed experimental information, including all training hyper-parameters and baseline implementations, is provided in the Appendix. 

\subsection{Graph-Level Tasks}
\label{sec:graph-task}
\paragraph{Datasets and Metrics.} For the graph-level task, we selected 13 real-world datasets from various sources, including seven datasets form the TU dataset (\textit{i.e.}, PROTEINS, NCI1, MUTAG, IMDB-B, IMDB-M, COLLAB, and REDDIT-B) and six from OGB (\textit{i.e.}, ogbg-molhiv, ogbg-moltox21, ogbg-moltoxcast, ogbg-molbbbp, ogbg-molesol, and ogbg-molfreesolv). Furthermore, these datasets comprise various domains ((\textit{i.e.}, biochemical, social network, and molecular) and tasks (\textit{i.e.}, classification and regression). For the aforementioned datasets, we strictly follow the evaluation metrics which are recommended by the given benchmarks. Specifically, for the TU dataset, we evaluated model performance based on the accuracy. For the OGB dataset, we used the area under the receiver operating characteristic curve (ROC-AUC) graph classification task, and the root mean squared error (RMSE) regression tasks. 

\paragraph{Results.} Table~\ref{tab:result-tu} presents the experimental results obtained from the TU dataset. The table demonstrates that DA-MoE can be integrated with three distinct GNN backbones, significantly enhancing their performance during graph-level tasks. Then, we also selected six OGB datasets to further demonstrate the depth-sensitivity phenomenon that appears in graph-structured datasets. The detailed results in Table~\ref{tab:result-ogb} offer insightful observations into DA-MoE's performance: \textbf{1) State-of-the-art performance:} The observation reveals significant performance improvements with DA-MoE, especially when utilizing GIN and GCN as backbone networks. Specifically, DA-MoE demonstrated consistent enhancements across all datasets, with a particularly notable increase of 9.06\% on the ogbg-molfreesolv and 2.72\% on IMDB-MULTI datasets with the GIN backbone. \textbf{2) Comparison with the MoE model:} Compared to GMoE~\cite{gmoe} which adopts the fixed GNN layer method, DA-MoE demonstrated formidable performance capabilities, further highlighting the depth-sensitivity issue present in graph datasets. Particularly, on the ogbg-moltoxcast dataset, GMoE model's performance declined compared to GIN backbone, whereas DA-MoE showed an improvement of 1.97\%.

\begin{table}[!t]
\centering
\renewcommand\arraystretch{1.25} % 行间距
\setlength\tabcolsep{5pt} % 列间距
\resizebox{0.4\textwidth}{!}{%
\begin{tabular}{@{}lcc@{}}
\toprule
    & \textbf{ogbn-proteins}              & \textbf{ogbn-arxiv}            \\ 
    & ROC-AUC & Accuracy   \\
    \midrule
GCN              
& $73.53_{\pm 0.56}$ & $71.74_{\pm 0.29}$      \\
\quad + GMoE             
& $74.48_{\pm 0.58} \ \footnotesize \begin{tcolorbox}[bluebox, left=-1.5pt, baseline=0.1cm]{\text{$\uparrow 1.29\%$}} \end{tcolorbox}$ 
& $71.88_{\pm 0.32} \ \footnotesize \begin{tcolorbox}[bluebox, left=-1.5pt, baseline=0.1cm]{\text{$\uparrow 0.20\%$}} \end{tcolorbox}$  \\
\quad + DA-MoE 
& $\textbf{75.22}_{\pm \textbf{0.85}} \ \footnotesize \begin{tcolorbox}[pinkbox, left=-1.5pt, baseline=0.1cm]{\text{$\uparrow \textbf{2.30\%}$}} \end{tcolorbox}$ 
& $\textbf{71.96}_{\pm \textbf{0.16}} \ \footnotesize \begin{tcolorbox}[pinkbox, left=-1.5pt, baseline=0.1cm]{\text{$\uparrow \textbf{0.31\%}$}} \end{tcolorbox}$  \\ \bottomrule
\end{tabular}%
}
\caption{\textbf{OGB node classification} experiments results on two datasets. The metrics are listed below each datasets, where higher scores indicate better performance. }
\label{tab:result-node}
\end{table}

\begin{table}[!t]
\centering
\renewcommand\arraystretch{1.25} % 行间距
\setlength\tabcolsep{5pt} % 列间距
\resizebox{0.4\textwidth}{!}{%
\begin{tabular}{@{}lcc@{}}
\toprule
     & \textbf{ogbl-ppa}          & \textbf{ogbl-ddi}           \\ 
    & HITS@100  & HITS@20  \\
    \midrule
GCN             
& $18.67_{\pm 1.32}$ & $37.07_{\pm 5.07}$      \\
\quad + GMoE     
& $19.25_{\pm 1.67} \ \footnotesize \begin{tcolorbox}[bluebox, left=-1.5pt, baseline=0.1cm]{\text{$\uparrow 3.11\%$}} \end{tcolorbox}$ 
& $37.96_{\pm 8.20} \ \footnotesize \begin{tcolorbox}[bluebox, left=-1.5pt, baseline=0.1cm]{\text{$\uparrow 2.40\%$}} \end{tcolorbox}$      \\
\quad + DA-MoE 
& $\textbf{21.46}_{\pm \textbf{2.53}} \ \footnotesize \begin{tcolorbox}[pinkbox, left=-2.5pt, baseline=0.1cm]{\text{$\uparrow \textbf{14.94\%}$}} \end{tcolorbox}$ 
& $\textbf{45.58}_{\pm \textbf{6.93}} \ \footnotesize \begin{tcolorbox}[pinkbox, left=-2.5pt, baseline=0.1cm]{\text{$\uparrow \textbf{22.96\%}$}} \end{tcolorbox}$  \\ 
\bottomrule
\end{tabular}%
}
\caption{\textbf{OGB link prediction} experiments results on two datasets. The metrics used are Hit rate (HITS@N), where higher scores indicate better performance. }
\label{tab:result-link}
\end{table}
\subsection{Node-Level Task}
\label{sec:node-task}
%In aggregating node information, using a single-scale aggregation strategy is suboptimal. Due to varying node degrees, high-degree nodes may receive excessive information, introducing noise, while low-degree nodes might not gather enough information at the same scale. Specifically, DA-MoE enables each node to select specific expert models, thereby easily transferring them to the node's task.
\paragraph{Datasets and Metrics.} We utilized two OGB datasets (\textit{i.e.}, ogbn-proteins and ogbn-arxiv) for the node classification task. Following the previous study~\cite{ogb-dataset,gmoe}, we used ROC-AUC as the evaluation metric for ogbn-proteins, and accuracy for ogbn-arxiv. 
\paragraph{Results.} Table~\ref{tab:result-node} shows the experimental results from the node classification task. We observed that DA-MoE improved significantly across all datasets when compared with the single expert GIN and fixed GNN layer GMoE. Notably, DA-MoE enhanced the ROC-AUC metric by 2.30\% and accuracy by 0.31\% on the ogbn-proteins and ogbn-arxiv datasets. This indicates that DA-MoE allows the nodes to adaptively select aggregation scales based on their degree, thereby enhancing model's performance.

\subsection{Link-Level Task}
\label{sec:link-task}
%For link prediction tasks, the logits of a link are multiplied by the representations of the corresponding pair of nodes. Therefore, we employ the same strategy used in node-level tasks to obtain high-quality node representations, thereby improving the performance of link prediction.
\paragraph{Datasets and Metrics.} We selected two datasets from OGB (\textit{i.e.}, ogbl-ppa and ogbl-ddi) for the link prediction task, and evaluated the model's performance based on HITS@N following the settings in OGB~\cite{ogb-dataset}. This measures the ratio of positive samples ranked among the top N against negative samples. Specifically, the HITS@100 and HITS@20 metrics were used for the ogbl-ppa and ogbl-ddi datasets, respectively.
\paragraph{Results.} Table~\ref{tab:result-link} summarizes the results, and we observed that the DA-MoE model's performance surpassed all baselines on these two datasets. Notably, DA-MoE outperformed the GIN backbone, delivering a substantial improvement of 14.94\% on ogbl-ppa and a remarkable 22.96\% on ogbl-ddi. Furthermore, when compared with other MoE models, DA-MoE's performance also improved significantly. The significant improvement over various baselines further demonstrate DA-MoE can effectively capture the nuances of different aggregation scales. The consistency of these enhancements across a variety of tasks and datasets underscores the model's excellent generalization capabilities.

\begin{figure}[!t] % !htb
\begin{center}
\includegraphics[width=1.0\linewidth]{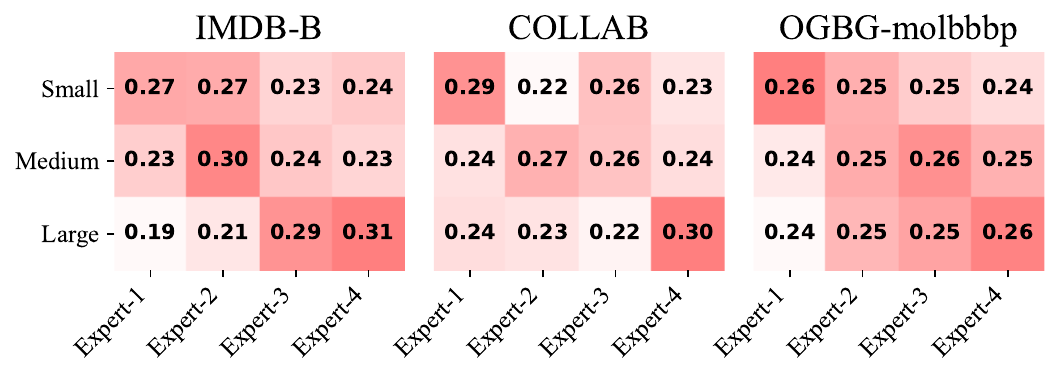}
\end{center}
\caption{Visualization of expert scores in the gating network. We report the mean scores for each expert with a specific graph scale determined by number of nodes. The rows indicate the scale of the graph, while the columns, labeled from ``Expert-1" to ``Expert-4", represent progressively deeper GNN layers associated with each expert. 
}
\label{fig:heatmap}
\end{figure}
\subsection{Visualization}
\label{sec:visualize}
To further explore the DA-MoE model's depth-sensitivity capacity, we visualized the gating network's scores across varying graph scales. Figure~\ref{fig:heatmap} illustrates the mean scoring results for each expert, and we obtain the following observations: \textbf{1) Depth-Sensitivity:} In the same dataset, smaller graphs tend to select shallow GNN layers. Meanwhile, large graphs prefer to use deep GNN layers to capture further neighbors' information. The experimental results further confirm depth-sensitivity characteristics in graph datasets. \textbf{2) Sparse Model:} We discovered that the distribution of expert scores is uniform, and sparse experts help focus on specific data scales, thereby improving the feature capturing process. This approach improves the model's overall performance while reducing operational costs.

\begin{figure}[!t] % !htb
\begin{center}
\includegraphics[width=1.0\linewidth]{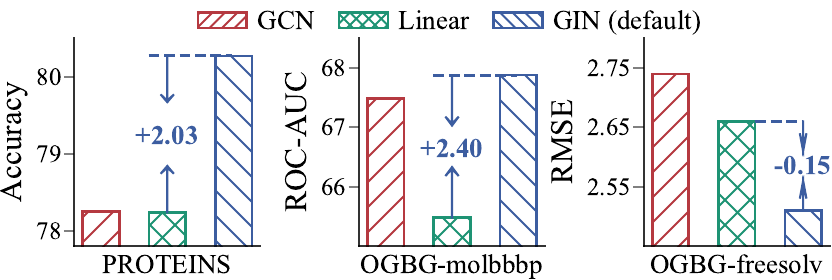}
\end{center}
\caption{The ablation study on gating network.
}
\label{fig:ablation}
\end{figure}

\subsection{Impact of Structure-Based Gating Network}
\label{sec:ablation-study}
As detailed before, our gating network utilizes a GNN model which consisting structural information instead of linear projection. Figure~\ref{fig:ablation} displays the study result to further evaluate the effectiveness of structure-based gating network. Based on the figure, we observe that combining the gating network with the GNN backbone consistently outperformed the model with linear projection. Specifically, the observed performance reduction in the absence of the structure-based gating network (\textit{i.e.}, 1.24\% on PROTEINS, 3.92\% on NCI1, and 0.47\% on IMDB-M). In summary, this modification greatly enhances the model's ability to capture the relationships and patterns within the gating network.

\begin{table}[!t]
\centering
\renewcommand\arraystretch{1.2} % 行间距
\setlength\tabcolsep{4pt} % 列间距
\resizebox{0.46\textwidth}{!}{%
\begin{tabular}{@{}l|cccccc@{}}
\toprule
 & \multicolumn{2}{c}{\textbf{ogbg-moltoxcast}} & \multicolumn{2}{c}{\textbf{ogbg-molbbbp}} \\ \cmidrule(lr){2-3} \cmidrule(lr){4-5}  
                & Time (ms) & Memory (MB) & Time (ms) & Memory (MB) \\ \midrule
GIN & 616.3 & 820 & 23.9 & 862      \\
\quad + GMoE & 2,247.9 & 1,030 & 64.8 & 1,076\\
\quad + DA-MoE & 643.6 & 970 & 30.3 & 1,020\\
\midrule
GCN  & 625.2 & 704 & 22.8 & 736       \\ 
\quad + GMoE   & 2,095.1 & 768 & 67.2 & 816\\
\quad + DA-MoE & 653.7 & 760 & 30.6 & 806\\
\bottomrule
\end{tabular}%
}
\caption{Detailed time \& memory of DA-MoE and other baseline methods on two datasets.}
\label{tab:time_cost}
\end{table}

\subsection{Efficiency Analysis}
Table~\ref{tab:time_cost} presents the running time and GPU memory usage during the inference process for the OGBG-moltoxcast and OGBG-molbbbp datasets, comparing them to two GNN backbone(\textit{i.e.}, GIN~\cite{gin} and GCN~\cite{gcn}) and one MoE model (\textit{i.e.}, GMoE~\cite{gmoe}). From the results, we observe two important insights: \textbf{1) Time Consumption:} DA-MoE displays an increase in time usage compared to each GNN backbone, but it is significantly lower than other MoE methods. This phenomenon can be explained by the fact that DA-MoE requires the computation of representations obtained from multiple GNN layers, which inevitably leads to higher time consumption. \textbf{2) Memory Usage:} DA-MoE has a slightly higher GPU memory cost compared to the backbone model, but is slightly lower than GMoE. This increase is primarily due to the storage of embeddings from various GNN layers, which necessitates additional memory for storing and processing these diverse representations. In summary, although DA-MoE incurs increased time and memory usage compared with backbones, these costs remain manageable and lead to significant performance improvements.

\begin{table}[!t]
\centering
\renewcommand\arraystretch{1.25} % 行间距
\setlength\tabcolsep{5pt} % 列间距
\resizebox{0.45\textwidth}{!}{%
\begin{tabular}{@{}lccc@{}}
\toprule
     & \textbf{PROTEINS}          & \textbf{ogbg-molbbbp}      & \textbf{ogbg-freesolv}       \\ 
     & Accuracy $\uparrow$ & AUC-ROC $\uparrow$ & RMSE $\downarrow$\\
    \midrule
GIN & $79.91_{\pm 0.62}$ & $65.50_{\pm 1.80}$ & $2.76_{\pm 0.35}$ \\
\quad + Dense Experts & $77.65_{\pm 0.62}$ & $66.93_{\pm 1.72}$ & $2.66_{\pm 0.32}$ \\
\quad + Sparse Experts& $\textbf{80.27}_{\pm \textbf{0.39}}$ & $\textbf{67.88}_{\pm \textbf{1.58}}$ & $\textbf{2.51}_{\pm \textbf{0.19}}$ \\
\bottomrule
\end{tabular}%
}
\caption{The parameter analysis of number of selected experts. The best results are in \textbf{bold}. ``Dense Experts" refers to the selected experts being equal total. ``Sparse Experts" denotes that the selected experts are fewer than total.}
\label{tab:para_k}
\end{table}
\begin{figure}[!t] % !htb
\begin{center}
\includegraphics[width=1.0\linewidth]{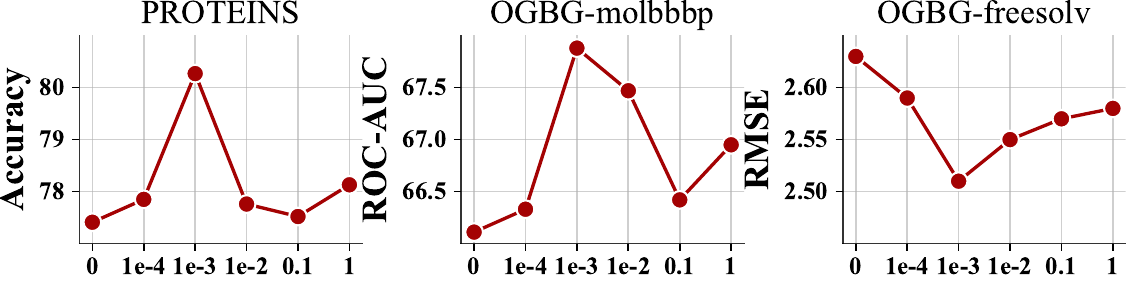}
\end{center}
\caption{The parameter analysis of scale factor $\lambda$.
}
\label{fig:lambda}
\end{figure}
\subsection{Parameter Analysis}
\label{sec:parameter-ans}
\paragraph{Number of Selected Experts.} The varying number of top experts selection (\textit{i.e.}, $k$) enables the model to adaptively capture the patterns from various GNN layers. As shown in Table~\ref{tab:para_k}, we observed that the optimal performance on these datasets is attained when the selected experts are sparse compared to the total. Specifically, compare to a dense model, a notable increase in the performance metrics by 3.37\%, 1.42\%, and 5.64\% was observed on the three datasets, respectively. This improvement underscores that sparse experts are capable of precisely capturing the aggregation information at different layers in GNNs, resulting in improved generalization. 
\paragraph{Balanced Loss Scale Factor.} In this section, we comprehensively analyze the impact of scale factor in balanced loss. As illustrated in Figure~\ref{fig:lambda}, we observed that the metric initially increases as lambda rises, peaking around the scale factor at $0.001$. Beyond this point, the model's performance gradually decreases. Notably, incorporating the balanced loss significantly impacted on the results, yielding a 3.69\%, 2.68\%, and 4.56\% improvement on the three datasets' performance. Despite the adjustments to the parameter $\lambda$ having a certain impact on the model's performance, the results consistently outperform the scenario where $\lambda=0$. This further confirms the effectiveness of the balanced loss approach.

\section{Conclusion} 
This study proposed DA-MoE, a novel MoE framework dedicated to addressing depth-sensitivity issue in graph-structured data. DA-MoE utilized different GNN layers as experts and allowed each individual graph to adaptively select experts. Additionally, this framework highlights two key modifications: the structure-based gating network and balanced loss function. Through comprehensive experiments covering graph, node, and link-level tasks, DA-MoE demonstrated a strong generalization capacity for varying dataset scales. Despite its competitive performance, DA-MoE still has room for improvement in certain areas: 1) Replacing the GNN backbone with the more powerful graph transformer, which may improve the model's representation capabilities. 2) The DA-MoE model may be expanded to graph self-supervised learning (\textit{e.g.}, contrastive learning).
\bibliography{aaai25}

\end{document}